\documentclass[preprint,12pt]{elsarticle}

\usepackage{amssymb}
\usepackage{amsmath}
\usepackage{graphicx}
\usepackage{multirow}%
\usepackage{amsfonts}%
\usepackage{amsthm}%
\usepackage{mathrsfs}%
\usepackage[title]{appendix}%
\usepackage{xcolor}%
\usepackage{textcomp}%
\usepackage{manyfoot}%
\usepackage{booktabs}%
\usepackage{algorithm}%
\usepackage{algorithmicx}%
\usepackage{algpseudocode}%
\usepackage{listings}%
\usepackage{makecell}%
\usepackage[colorlinks,
            linkcolor=red, 
             anchorcolor=blue, 
            citecolor=green, 
            ]{hyperref}

\journal{Arxiv}

\begin{document}

\begin{frontmatter}



\title{Self-Paced Learning Strategy with Easy Sample Prior Based on Confidence for the Flying Bird Object Detection Model Training}


\author[inst1]{Zi-Wei Sun}
\author[inst1]{Ze-Xi hua \corref{label1}}
\author[inst1]{Heng-Chao Li}
\author[inst1]{Yan Li}

\affiliation[inst1]{organization={School of Information Science and Technology, Southwest Jiaotong University},
            city={Chengdu},
            postcode={610000}, 
            country={China}}
\cortext[label1]{Corresponding author at: School of Information Science and Technology, Southwest Jiaotong University, Chengdu 610000, China. Email address: xx\_zxhua@swjtu.edu.cn}

\begin{abstract}
In order to avoid the impact of hard samples on the training process of the Flying Bird Object Detection model (FBOD model, in our previous work, we designed the FBOD model according to the characteristics of flying bird objects in surveillance video), the Self-Paced Learning strategy with Easy Sample Prior Based on Confidence (SPL-ESP-BC), a new model training strategy, is proposed. Firstly, the loss-based Minimizer Function in Self-Paced Learning (SPL) is improved, and the confidence-based Minimizer Function is proposed, which makes it more suitable for one-class object detection tasks. Secondly, to give the model the ability to judge easy and hard samples at the early stage of training by using the SPL strategy, an SPL strategy with Easy Sample Prior (ESP) is proposed. The FBOD model is trained using the standard training strategy with easy samples first, then the SPL strategy with all samples is used to train it. Combining the strategy of the ESP and the Minimizer Function based on confidence, the SPL-ESP-BC model training strategy is proposed. Using this strategy to train the FBOD model can make it to learn the characteristics of the flying bird object in the surveillance video better, from easy to hard. The experimental results show that compared with the standard training strategy that does not distinguish between easy and hard samples, the $\text{AP}_{50}$ of the FBOD model trained by the SPL-ESP-BC is increased by 2.1\%, and compared with other loss-based SPL strategies, the FBOD model trained with SPL-ESP-BC strategy has the best comprehensive detection performance. This  project is publicly available at  \href{https://github.com/Ziwei89/FBOD-BSPL}{https://github.com/Ziwei89/FBOD-BSPL}.
\end{abstract}


\end{frontmatter}


\section{Introduction}\label{intro}

Detecting flying bird objects has important applications in many fields, such as repelling birds in airports \cite{shi_airport_repelling_birds, wu_airport_repelling_birds_a_new_skeleton}, preventing birds in crops \cite{zhao_preventing_birds_in_crops, Shivam_preventing_birds_in_crops}, avoiding bird collisions in wind power stations \cite{Yoshihashi_avoiding_bird_collisions_in_wind_power, Christopher_avoiding_bird_collisions_in_wind_power_Automated_monitoring_for_birds}, etc. We are working on using surveillance cameras to detect flying birds in real-time.

\begin{figure*}[!ht]
\centering
\includegraphics[width=5in]{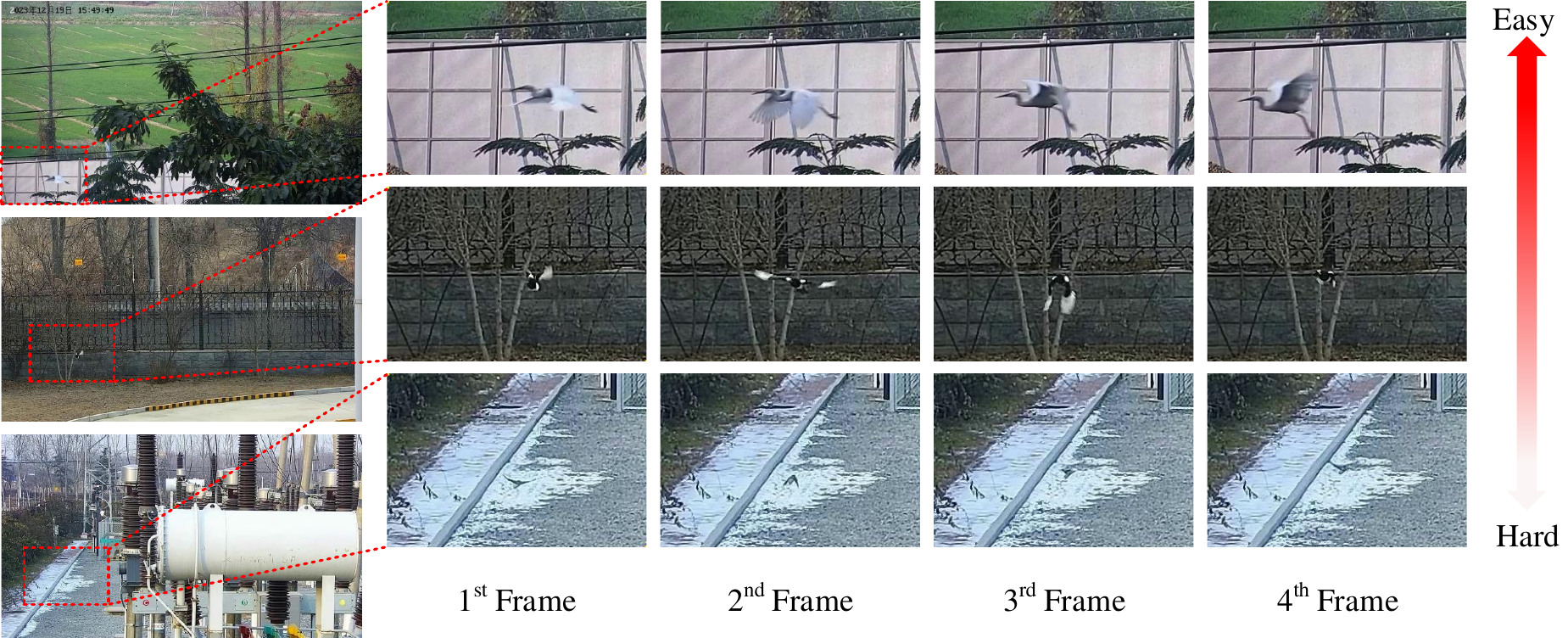}
\caption{Identifying the flying bird object in the surveillance video has different degrees of difficulty.}
\label{different_degrees_of_difficulty_fig}
\end{figure*}

The identification of flying birds in surveillance video has different difficulty attributes. Specifically, through manual observation, birds in some video clips can be easily identified using a single frame image. In some video clips, birds need to be identified by careful observation of a single-frame image. Single-frame images cannot identify some video clips, but birds can be easily identified by observing consecutive frames of images. In some video clips, birds can only be identified by carefully observing consecutive frames of images, as shown in Fig. \ref{different_degrees_of_difficulty_fig}. 

In deep learning, we usually divide the dataset into a training set used to train the model and a test set used to evaluate the model performance. However, this practice may encounter problems when dealing with samples with different recognition difficulties. Since there is no clear distinction between easy and hard samples, the model may be affected by noise in hard samples during training, leading to performance degradation. Take the flying bird object in the last row on the right in Fig. \ref{different_degrees_of_difficulty_fig} as an example: its similarity to the background may make it difficult for the model to distinguish it, thus misrecognizing the background as a flying bird object. To mitigate the influence of hard samples on the training process, a common strategy is to utilize only easy samples for training, thereby reducing false detections (analogous to removing noisy data in scenarios with noisy labels \cite{lu_mentornet, malach_when_to_update, Han_co_teacher, shen_learning_with_bad_training_data}). However, there is also a risk because the absence of hard samples may cause the model to perform poorly in the face of complex scenes, making it difficult to detect hard samples accurately.

There is a widely used method for hard samples, namely the Hard Example Mining (HEM) algorithm \cite{Girshick_HEM, he_HEM_focal_loss, li_HEM_Gradient_Harmonized}, which selects the most hard samples for training in each batch (or assigns higher weights to these hard samples). This kind of approach is suitable for cleaner datasets, as hard samples can provide additional information \cite{chang_hard_samples_more_information_Active_Bias}. However, the flying bird objects that need to be detected in this paper are mostly challenging due to noise (background). Such hard samples contain less information than easy samples (unambiguous objects), as exemplified by the flying bird object in the third row of Fig. \ref{different_degrees_of_difficulty_fig}. Suppose the HEM method is still employed to train the model under such circumstances. In that case, it will be difficult for the model to learn the characteristics of flying bird objects, and there is a risk of overfitting the hard samples, which is prone to increased false detections during the model inference stage.

There is also a class of machine learning paradigms that mimic the learning patterns of humans or animals, contrary to the methods of HEM. The core idea of this mode is that in the learning process, easy samples are introduced first, and then hard samples are introduced gradually.  By adopting this learning paradigm, the model's training can be guaranteed to converge stably and quickly in the presence of noisy or abnormal labels.  This learning paradigm is known as curriculum learning \cite{Bengio_Curriculum_Learning, wang_surver_Curriculum_Learning}.

The difficulty measurement and training scheduling of traditional predefined curriculum learning are all designed manually \cite{ionescu_How_Hard_Can_It_Be_cl_manually, chen_Webly_Supervised_Learning_cl_manually, soviany_Image_Difficulty_CuGAN_cl_manually, gui_Curriculum_Learning_for_Facial_Expression_cl_manually}, and there are many limitations, among which the difficulty measurement is the most difficult \cite{wang_surver_Curriculum_Learning}. Difficulty measures often require expert domain knowledge to design. However, examples that are easy for humans are not always easy for models because the decision boundaries of models and humans are fundamentally different \cite{yuan_Adversarial_Examples_model_different_from_man}. Based on the above limitations, automatic curriculum learning strategies have been widely proposed \cite{kumar_SPL_for_latent_variable, alex_auto_cl, wei_STC_auto_cl, jonathan_On_the_effectiveness_of_SPL_auto_cl, guy_On_The_Power_of_Curriculum_Learning_auto_cl, matiisen_teacher_student_curriculum_learning_auto_cl}; among them, Self-paced Learning (SPL) \cite{kumar_SPL_for_latent_variable} is a simple and widely used automatic curriculum learning algorithm. SPL uses the training loss as the automatic difficulty metric and then introduces a regularizer to automatically select the appropriate hard samples for learning according to their learning degree. Inspired by the SPL algorithm, this paper considers the idea of SPL to train the Flying Bird Object Detection (FBOD) model in surveillance video (In our previous work \cite{sun2024flying}, we designed the FBOD model according to the characteristics of flying birds in surveillance video), to cope with the situation that the recognition difficulty of the flying bird object in the surveillance video is different.

When the SPL is applied to the training of the FBOD model in surveillance video, this paper improves SPL and proposes the SPL strategy with Easy Sample Prior Based on Confidence (SPL-ESP-BC). Firstly, the loss-based Minimizer Function in SPL is improved, and the confidence-based Minimizer Function is proposed, which makes it more suitable for one-class object detection tasks. Secondly, to give the model the ability to judge easy and hard samples at the early stage of training by using the SPL strategy, an SPL strategy with Easy Sample Prior (ESP) is proposed. The FBOD model is trained using the standard training strategy with easy samples first, then the SPL strategy with all samples is used to train it. Combining the strategy of the ESP and the Minimizer Function based on confidence, the SPL-ESP-BC model training strategy is proposed. Using this strategy to train the FBOD model can make it to learn the characteristics of the flying bird object in the surveillance video easier, from easy to hard.

The main contributions of this paper are as follows.

\begin{enumerate}
\item{An SPL strategy Based on Confidence (SPL-BC) for one-class object detection model is proposed. The confidence-based Minimizer Function is used to determine the optimal weight of the sample in the SPL training process, simplifying the strategy of judging whether the sample is hard or not and making the SPL training process of the one-class object detection model simpler and more intuitive.}
\item{An SPL strategy with Easy Sample Prior Based on Confidence (SPL-ESP-BC) for the FBOD model training is proposed. Firstly, the manually selected easy flying bird object samples pre-train the FBOD model. Then, the manually selected easy samples are mixed with the overall samples. The SPL-BC strategy is used to retrain the FBOD model, which eliminates the subjective influence of the manual evaluation of the simplicity of the sample. At the same time, it avoids the problem of the initial model being unable to identify easy or hard samples and falling into the disordered search state.}
\item{Under the framework of SPL-ESP-BC, a confidence-based Minimizer Function example and a training schedule example are given. Based on the examples, a series of quantitative and qualitative experiments are designed to prove the effectiveness of the SPL-ESP-BC strategy for the FBOD model.}
\end{enumerate}

The remainder of this paper is structured as follows: Section \ref{related_works} presents work related to this paper. In Section \ref{spl-bc}, the SPL-BC strategy for one-class object detection model is described. Section \ref{spl-esp-bc} describes the SPL-ESP-BC strategy in detail. Section \ref{experiments} presents a comparative experiment of the proposed method. Section \ref{conclusion} concludes our work.

\section{Related Works}\label{related_works}

In the previous work \cite{sun2024flying}, we studied the characteristics of the flying bird object in surveillance video, such as unobvious features in single frame images, small size in most cases, and asymmetric rules, and proposed a FBOD method for Surveillance Video (FBOD-SV), which can detect the flying bird object in surveillance video. In this paper, based on the work \cite{sun2024flying}, we will adopt the SPL idea to deal with the problem of different degrees of difficulty in identifying flying birds in surveillance videos. To facilitate the introduction of the following contents, we briefly review the FBOD-SV method and the SPL algorithm in this subsection.

\subsection{The FBOD-SV Method}

\subsubsection{Overview of the FBOD-SV Method}

The FBOD-SV \cite{sun2024flying} primarily addresses the issues of feature loss during feature extraction due to the unclear features of flying birds in single-frame images of surveillance videos, the difficulty in detecting objects due to their small size in most cases, and the challenge of allocating positive and negative samples during model training caused by asymmetric shapes. Specifically, first, FBOD-SV employs a feature aggregation unit based on correlation attention to aggregate the features of flying birds across consecutive frames. Secondly, it adopts a network structure that first downsamples and then upsamples to fully integrate shallow and deep feature map information, utilizing a large feature layer of this network to predict flying birds with special multi-scales (mostly small scales) in surveillance videos. Finally, during the training process of the model, the SimOTA-OC dynamic label assignment method is utilized to handle the possible irregular shapes of flying bird objects in surveillance videos.

\subsubsection{The Loss Function in FBOD-SV}

FBOD-SV \cite{sun2024flying} belongs to the object detection method based on the anchor-free class, which does not use the preset anchor box and directly uses the feature points (anchor points) of the output feature map to predict the flying bird object. The FBOD-SV model has two output branches: the confidence prediction branch and the location regression branch. The confidence prediction branch is used to predict whether the anchor (point) sample\footnote{This paper deals with two types of samples, namely anchor samples and object samples. The anchor samples refer to the feature pixels of the feature map, while the object samples refer to flying bird objects, and a flying bird object can contain multiple anchor samples. In general, it is not specifically stated that the sample specifically refers to the object sample.} is positive (an anchor sample is a positive sample when it belongs to a bird object and a negative sample when it belongs to the background. It is difficult to determine which bird object the anchor sample belongs to or when it is at the edge of the object bounding box, it can be set to ignore the sample and not handle it in training), and the position regression branch is used to return the bounding box information of the bird object.

FBOD-SV uses a multi-task loss function to train the FBOD model, which includes a confidence loss and a position regression loss. Specifically, the loss of a certain anchor sample is expressed as the weighted sum of the confidence loss and the position regression loss as follows,
\begin{align}\label{eq:lai}
\text{L} \left( {\text{A}}_{i} \right ) = \text{L}_{\text{Conf}}\left ( {\text{A}}_{i} \right ) + \alpha \text{L}_{\text{Reg}}\left ( {\text{A}}_{i} \right ),
\end{align}
where $\text{L}_{\text{Conf}}\left ( \cdot \right )$ represents confidence loss and the L2 loss is used. $\text{L}_{\text{Reg}}\left ( \cdot \right )$ stands for position regression loss, and the CIOU \cite{zheng_CIOU} loss is adopted. $\alpha$ is the weighted balance parameter for the two losses. During training, the total loss is equal to the sum of all anchor losses, as follows,
\begin{align}\label{eq:tl}
\text{Total Loss} &=\frac{1}{\text{N}} \sum{\text{L}\left (\text{A}_i\right)}\notag\\
&=\frac{1}{\text{N}}\left({\sum{\text{L}}_{\text{Conf}}\left (\text{A}_i\right)} + {\alpha}\sum{{\text{L}}_{\text{Reg}}\left (\text{A}_i\right)}\right)\notag\\
&=\frac{1}{\text{N}}\left({\text{L}}_{\text{C}}+{\alpha}{\text{L}}_{\text{R}} \right),
\end{align}
where $\text{N}$ is a normalized parameter, when the image contains bird objects, the $\text{N}$ is the number of positive anchor samples; otherwise, the $\text{N}$ is a fixed positive number.  ${\text{L}}_{\text{C}}$ and  ${\text{L}}_{\text{R}}$ are the confidence loss and the position regression loss for all anchors, respectively.

\subsection{The SPL Algorithm}

Given a training dataset $\textbf{D} = \left\{ \left (\textbf{x}_i,\text{y}_i \right ) \right \}_{(i=1)}^n$, where $\textbf{x}_i$, $\text{y}_i$ denotes the $i^\text{th}$ input sample and its label, respectively. When the input is $\textbf{x}_i$, $f\left (\textbf{x}_i;\textbf{w} \right )$ represents the prediction result of the model $f$, where w is the model $f$'s parameters. $\text{L} \left (\text{y}_i,f\left (\textbf{x}_i;\textbf{w} \right ) \right )$ represents the loss between the predicted result and the label. Then, the expression of the SPL \cite{kumar_SPL_for_latent_variable} is as follows,
\begin{align}\label{eq:spl}
\min\limits_{\textbf{w}, \textbf{v}} \text{E} \left( \textbf{w}, \textbf{v}, \lambda \right) = \sum\limits_{i=1}^n \left( v_i \text{L} \left (\text{y}_i,f\left (\textbf{x}_i;\textbf{w} \right ) \right ) + g\left( v_i, \lambda \right) \right),
\end{align}
where $\lambda$ is the age parameter that controls the learning speed, $\textbf{v}=[v_1, v_2, \cdots, v_n]$ is the sample weight used to determine which samples participate in the training or the degree of participation. $g\left( v_i, \lambda \right)$ denotes Self-Paced Regularizer. As the age parameter $\lambda$ increases, the samples to be learned can be gradually included in the training from easy to hard by optimizing the algorithm and alternately fixing one of $\textbf{w}$ and $\textbf{v}$ to optimize the other. Specifically, when the sample weight parameter $\textbf{v}$ is given, the optimal parameter $\textbf{w}$ of the model is given by the Weighted Loss Function,
\begin{align}\label{eq:splw}
\min\limits_{\textbf{w}} \sum\limits_{i=1}^n  v_i \text{L} \left (\text{y}_i,f\left (\textbf{x}_i;\textbf{w} \right ) \right ).
\end{align}
When the model weight $\textbf{w}$ is given, the optimal sample weight $\textbf{v}$ is determined by the optimization formula as follows,
\begin{align}\label{eq:splv}
\min\limits_{\textbf{v}} \sum\limits_{i=1}^n \left( v_i \text{L} \left (\text{y}_i,f\left (\textbf{x}_i;\textbf{w} \right ) \right ) + g\left( v_i, \lambda \right) \right).
\end{align}
When calculating the optimal sample weight parameter $\textbf{v}$, since the model weight $\textbf{w}$ is fixed, the loss of the $i^\text{th}$ sample is a constant, so the optimal value of the weight $v_i$ is uniquely determined by the corresponding Minimizer Function $\sigma\left( \lambda, \text{L} \left (\text{y}_i,f\left (\textbf{x}_i;\textbf{w} \right ) \right ) \right)$, and has
\begin{align}\label{eq:spl_minimizer}
\sigma\left( \lambda, l_i \right) l_i + g\left( \sigma\left( \lambda, l_i \right), \lambda \right) \leq v_i l_i + g\left( v_i, \lambda \right), \forall v_i \in [0, 1],
\end{align}
where $l_i = \text{L} \left (\text{y}_i,f\left (\textbf{x}_i;\textbf{w} \right ) \right )$.

If $g\left( v_i, \lambda \right)$ has a concrete analytical form, it is called Self-Paced Explicit Regularizer. Table \ref{tab:Minimizer_Functions} shows some classical Self-Paced Explicit Regularizers, together with closed-form solutions (Minimizer Functions) for the optimal weights.
\begin{table}[!ht]
\caption{Some classical  Self-Paced Explicit Regularizers and their closed-form solutions.\label{tab:Minimizer_Functions}}
\centering
\begin{tabular}{c|c|c}
\hline
Names & Regularizers & Closed-form solutions \\
\hline
Hard \cite{kumar_SPL_for_latent_variable} & $-\lambda \sum\limits_{i=1}^n v_i$ & $\begin{cases} 1, &l_i < \lambda,\\0, &{\text { Otherwise}},\end{cases}$ \\
\hline
Linear \cite{jiang_easy_samples_first_sp_regularizers} & $\frac{1}{2} \lambda \sum\limits_{i=1}^n\left( v_i^2 - 2 v_i\right) $ & $\begin{cases} 1 - l_i/\lambda, &l_i < \lambda,\\0, &{\text { Otherwise}},\end{cases}$ \\
\hline
Logarithmic \cite{jiang_easy_samples_first_sp_regularizers} & $\makecell[c]{\sum\limits_{i=1}^n\left( \zeta v_i - \frac{\zeta ^{v_i}}{\text{log} \zeta}\right)\\\zeta = 1 - \lambda, 0 < \lambda < 1}$ & $\begin{cases} \frac{\text{log}\left( l_i + \zeta \right)}{\text{log} \zeta}, &l_i < \lambda,\\0, &{\text { Otherwise}},\end{cases}$ \\
\hline
\end{tabular}
\end{table}

Fan et al. \cite{Yanbo_SPL_implicit_regularization} further proposed the Self-Paced Implicit Regularizer (refer to \cite{Yanbo_SPL_implicit_regularization} for the definition of the Self-Paced Implicit Regularizer). At the same time, Fan et al. also proposed an SPL framework based on a Self-Paced Implicit Regularizer named SPL-IR \cite{Yanbo_SPL_implicit_regularization},
\begin{align}\label{eq:spl_implicit}
\min\limits_{\textbf{w}, \textbf{v}} \text{E} \left( \textbf{w}, \textbf{v}, \lambda \right) = \sum\limits_{i=1}^n \left( v_i \text{L} \left (\text{y}_i,f\left (\textbf{x}_i;\textbf{w} \right ) \right ) + \psi \left( v_i, \lambda \right) \right),
\end{align}
where $\psi \left( v_i, \lambda \right)$ is the Self-Paced Implicit Regularizer. A selective optimization algorithm can solve Eq. \eqref{eq:spl_implicit}. Different from ordinary SPL, the analytical form of $\psi \left( v_i, \lambda \right)$ in the Self-Paced Implicit Regularizer in Eq. \eqref{eq:spl_implicit} can be unknown, and the optimal weight $v^*$ is determined by the Minimizer Function $\sigma (\lambda,l_i)$.

\section{The SPL-BC Strategy for One-class Object Detection Model}\label{spl-bc}

In this paper, FBOD belongs to one-class object detection. When training a one-class object detection model, its loss function does not have class loss. Based on the principle of SPL-IR \cite{Yanbo_SPL_implicit_regularization}, we deduce that one-class object detection can use the prediction confidence of the model to determine the optimal weight of the Weighted Loss Function in SPL. The specific derivation process is as follows.

When training the object detection model using SPL strategy, the loss function can be expressed as follows,
\begin{align}\label{eq:lod}
L = L^\text{neg} + \sum\limits_{i=1}^n v_i l_i^\text{pos},
\end{align}
where $L^\text{neg}$ represents the negative sample loss, $l_i^\text{pos}$ represents the loss of the $i^\text{th}$ positive sample, and $v_i$ represents the weight corresponding to the $i^\text{th}$ positive sample. This weight determines whether (or to what extent) the corresponding sample is involved in training. This weight is related to the difficulty of the object; the harder the object is, the smaller the corresponding weight value (or 0), indicating that the hard object is less involved in the training (or not involved in the training). This loss function is called the Weighted Loss Function.

In the SPL-IR framework \cite{Yanbo_SPL_implicit_regularization}, the optimal weights of the Weighted Loss Function are determined by the Minimizer Function without knowing the analytical form of the Self-Paced Implicit Regularizer. For example, the optimal weight corresponding to the $i^\text{th}$ positive sample is
\begin{align}\label{eq:minimizer_1}
v_i^* = \sigma (\lambda,l_i^\text{pos}),
\end{align}
where the loss of $l_i^\text{pos}$ can be viewed as the difficulty value of the $i^\text{th}$ positive sample. In the one-class object detection task, the loss of positive samples does not include the class loss, but only the confidence and position regression loss,
\begin{align}\label{eq:l_pos}
l_i^\text{pos} = l_{i\text{conf}}^\text{pos} + \alpha l_{i\text{reg}}^\text{pos},
\end{align}
where $l_{i\text{conf}}^\text{pos}$ and  $l_{i\text{reg}}^\text{pos}$ is the confidence loss and the position regression loss of the $i^\text{th}$ positive sample respectively, and $\alpha$ is the equilibrium parameter of the two losses. If the confidence loss of the $i^\text{th}$ positive example is large (prediction confidence is small), the position regression loss is small, and the total loss is small, the sample may be classified as easy. However, in the inference prediction stage, the object with small confidence is not easily recognized, even if its total loss value is small. Therefore, in one-class object detection, using the total loss value of samples to measure whether it is hard is inaccurate, and it is more reasonable to use prediction confidence. Therefore, the Minimizer Function can be designed using only the confidence loss of the samples,
\begin{align}\label{eq:minimizer_2}
v_i^* = \sigma (\lambda,l_{i\text{conf}}^\text{pos}).
\end{align}
The confidence loss can be expressed as a function of the distance between the predicted confidence and the GT value ``1'', 
\begin{align}\label{eq:l_pos_conf}
l_{i\text{conf}}^\text{pos} = \text{func}(|\text{Conf}_{\text{pred}}(i) - 1|),
\end{align}
where $\text{Conf}_{\text{pred}}(i)$ is the prediction confidence of the $i^\text{th}$ positive sample, which ranges from 0 to 1, and $\text{func} \left(\cdot \right)$ represents some kind of function mapping. Substituting Eq. \eqref{eq:l_pos_conf} into Eq. \eqref{eq:minimizer_2} gives
\begin{align}\label{eq:minimizer_3}
v_i^* = \sigma (\lambda,\text{func} (|\text{Conf}_{\text{pred}}(i) - 1|)),
\end{align}
where $\lambda$ can be understood as the threshold parameter of hard samples (the threshold related to sample loss), which gradually increases with the number of training iterations, indicating that hard samples (samples with large loss) gradually participate in training as training proceeds. The large confidence loss of hard samples is equivalent to the small prediction confidence. Let $\lambda = \varrho ( \xi )$, $\xi$ is inversely correlated with $\lambda$, then $\xi$ gradually decreases with the increase of training iterations, which can indicate that hard samples (samples with smaller prediction confidence) gradually participate in training as training proceeds. Therefore, $\xi$ can also be understood as the threshold parameter of hard samples (the threshold related to the prediction confidence of the sample). Substituting  $\lambda = \varrho ( \xi )$ into Eq. \eqref{eq:minimizer_3} gives,
\begin{align}\label{eq:minimizer_4}
v_i^* = \sigma (\varrho ( \xi ),\text{func} (|\text{Conf}_{\text{pred}}(i) - 1|)).
\end{align}
Eq. \eqref{eq:minimizer_4} shows that the Minimizer Function to determine the optimal weight of a sample can be expressed as a function related to the parameter $\xi$ and the prediction confidence $\text{Conf}_{\text{pred}}(i)$ for this sample. We simplify Eq. \eqref{eq:minimizer_4} to obtain the confidence-based Minimizer Function, 
\begin{align}\label{eq:minimizer_5}
v_i^* = \sigma' (\xi, \text{Conf}_{\text{pred}}(i)).
\end{align}
When the prediction confidence of a sample is close to 1, it can be said that the sample is easy. When the confidence goes to 0, we can indicate that the sample is hard. Parameter $\xi$ varies from large to small between [0, 1], which means that the hard samples (the samples with low prediction confidence) are gradually involved in the training as the training progresses. The setting of the hyperparameters of the Minimizer Function based on confidence is simple and intuitive.

The sample weight of the Weighted Loss Function is fixed, and the model's weight is optimized through Eq. \eqref{eq:minimizer_1}. After that, the model's weight is fixed, and the optimal weight of the Weighted Loss Function is determined through Eq. \eqref{eq:minimizer_5}. In this way, the model weight and sample weight are optimized alternately and iteratively, which is the SPL strategy Based on Confidence (SPL-BC) for one-class object detection model.

\section{The FBOD model training strategy based on SPL-ESP-BC}\label{spl-esp-bc}

\begin{figure*}[!ht]
\centering
\includegraphics[width=4.5in]{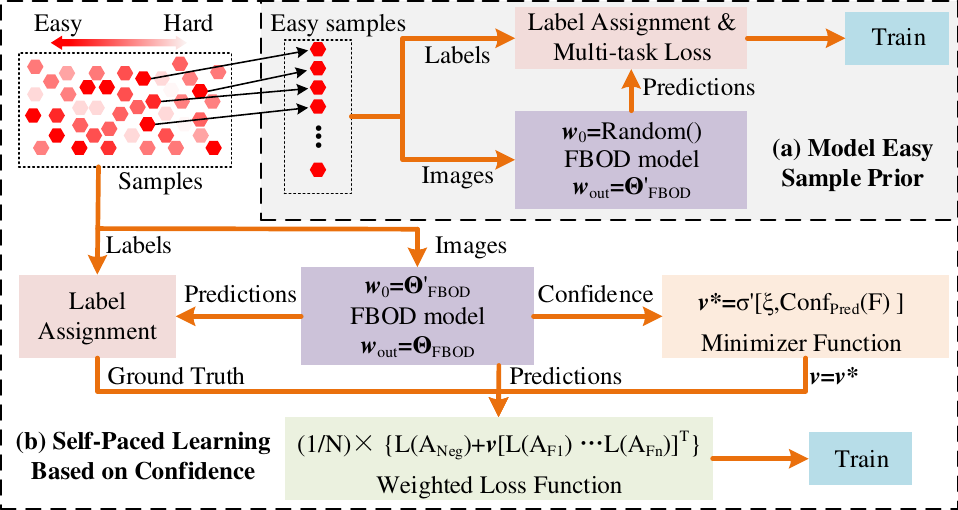}
\caption{The FBOD model training strategy based on SPL-ESP-BC.}
\label{method_fig}
\end{figure*}

Fig. \ref{method_fig} shows the block diagram of the FBOD model training strategy based on SPL-ESP-BC proposed in this paper. There are two main parts: model Easy Sample Prior (ESP) and Self-Paced Learning Based on Confidence (SPL-BC). Specifically, firstly, easy samples are used to train the FBOD model (The weights of the model are initialized with random numbers.), so that it has the ability to recognize easy and hard samples [as shown in Fig. \ref{method_fig}(a)]. Then, the SPL-BC strategy is used to train the FBOD model (the model weights are initialized using the model weights after training the model with easy samples). The model prediction confidence is input into Minimizer Function to determine the optimal weight of the Weighted Loss Function in the SPL, and then control which samples do not participate in the training and which samples participate in the training (or the degree of participation) [as shown in Fig. \ref{method_fig}(b)].

Next, the necessary Weighted Loss Function and Minimizer Function when the SPL-BC is applied to the training of the FBOD model are introduced first. Then, the FBOD model training strategy based on SPL-ESP-BC is described in detail.

\subsection{The Weighted Loss Function and Minimizer Function When Applying the SPL-BC to the FBOD Model}\label{step1}

As know from Section \ref{spl-bc}, to apply the SPL-BC strategy, two types of functions need to be determined: the Weighted Loss Function that optimizes the model weights and the Minimizer Function that determines the sample weights. Next, we will introduce the Weighted Loss Function and the Minimizer Function when applying the SPL-BC training strategy to the FBOD (one-class object detection) model.

\subsubsection{The Weighted Loss Function When Applying the SPL-BC to the FBOD Model}\label{step1_1}
The anchor samples are divided into negative samples, positive samples of flying bird object 1, positive samples of flying bird object 2, ..., and positive samples of the bird object $n$. Since the loss of negligible samples is always 0, Eq. \eqref{eq:tl} can be rewritten as follows,
\begin{align}\label{eq:tl_2}
\text{Total Loss} =\frac{1}{\text{N}} \left (\text{L}\left (\text{A}_{neg}\right) + \text{L}\left (\text{A}_{F_1}\right) + ... + \text{L}\left (\text{A}_{F_n}\right)\right),
\end{align}
where $\text{A}_{neg}$ represents the negative anchor sample set, $\text{A}_{F_i} (i \in (1, ... ,n))$ denotes the set of positive anchor samples of the bird object $i$. When training a model using SPL strategy, the Weighted Loss Function can be expressed as follows,
\begin{align}\label{eq:wlf}
\text{Total Loss} &=\frac{1}{\text{N}} \left (\text{L}\left (\text{A}_{neg}\right) + v_i \text{L}\left (\text{A}_{F_1}\right) + ... + v_n \text{L}\left (\text{A}_{F_n}\right)\right)\notag\\
&=\frac{1}{\text{N}} \left (\text{L}\left (\text{A}_{neg}\right) + \boldsymbol{\Vec{v}} \left( \text{L}\left (\text{A}_{F_1}\right) ...\ \text{L}\left (\text{A}_{F_n}\right)\right)^T \right),
\end{align}
where $\boldsymbol{\Vec{v}}=[v_1\ ...\ v_n]$ is the weight corresponding to the object sample loss, which controls which bird objects participate in the training or the degree of participation in the training.

\subsubsection{The Minimizer Function When Applying the SPL-BC to the FBOD Model}\label{step1_2}

\begin{figure*}[!ht]
\centering
\includegraphics[width=4in]{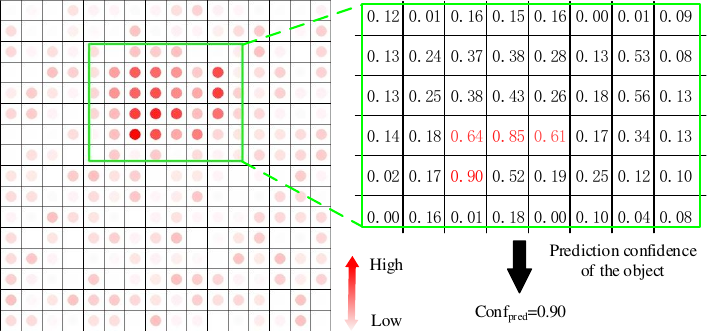}
\caption{Illustration of calculating the prediction confidence for a bird object.}
\label{conf_pred_fig}
\end{figure*}

According to Eq. \eqref{eq:minimizer_5}, the Minimizer Function required by the SPL-BC strategy for the FBOD model can be directly given as follows,
\begin{align}\label{eq:mf}
v_i^* = \sigma' (\xi, \text{Conf}_{\text{pred}}(F_i)),
\end{align}
where $\text{Conf}_{\text{pred}}(F_i)$ represents the prediction confidence of the bird object $i$. Each anchor in the confidence prediction feature map has a confidence prediction value, and each bird object has multiple anchor points, so each bird object has multiple confidence predictions. Similar to the way of calculating the confidence of the flying bird object when detecting the flying bird object, this paper takes the maximum prediction value as the confidence prediction value of the flying bird object as follows, 
\begin{align}\label{eq:conf_pred}
\text{Conf}_{\text{pred}}(F_i) = \text{max}_\text{conf} (\text{box}_\text{gt}(F_i), \text{Conf}_{\text{pred}}),
\end{align}
among them, $\text{box}_\text{gt}(F_i)$ represents the GT bounding box of the bird object $i$, $\text{Conf}_{\text{pred}}$ represents the confidence prediction feature map, and $\text{max}_\text{conf}(\cdot)$ represents the calculation of the prediction confidence value of the bird object [Fig. \ref{conf_pred_fig} shows the schematic diagram of the calculation process. The green box represents the GT bounding box of a certain bird object. The left of Fig. \ref{conf_pred_fig} shows the confidence output feature map, where the depth of the point color represents the magnitude of the prediction confidence of the feature point (anchor point). The right of Fig. \ref{conf_pred_fig} shows the predicted confidence values of all anchors of the bird object, where the maximum predicted confidence is the predicted confidence of the bird object].

This paper uses a piecewise function on prediction confidence as an example of the Minimizer Function. Specifically, when the difficulty of the bird object sample is less than a certain threshold (the prediction confidence is greater than a certain threshold), the sample weight value is determined by the root of the prediction confidence. Otherwise, the sample will not participate in the training (sample weight is 0). The Minimizer Function is as follows,
\begin{align}\label{eq:mf_example}
v_i=
\begin{cases}
\sqrt[m]{\text{Conf}_{\text{pred}}(F_i)}, &\text{Conf}_{\text{pred}}(F_i) > \xi,\\
0, &{\text { Otherwise}},
\end{cases}
\end{align}
where $m$ is a positive integer (in the subsequent experiments, $m$ is set to 3). The qualitative interpretation of Eq. \eqref{eq:mf_example} is as follows: when the difficulty of a flying bird object exceeds a certain threshold, it will not participate in the training (the weight corresponding to the loss is 0). When a flying bird object is easier to recognize [the prediction confidence $\text{Conf}_{\text{pred}}(F_i)$ is larger], its participation in training is stronger.

\begin{figure*}[!ht]
\centering
\includegraphics[width=2.5in]{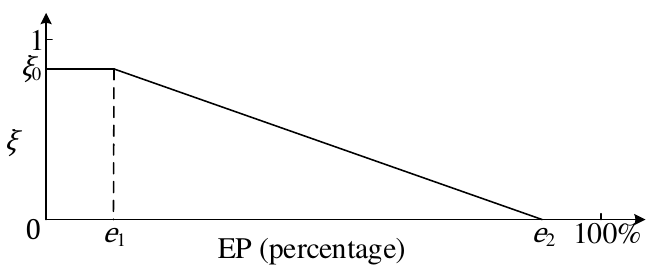}
\caption{The relationship between $\xi$ and training process.}
\label{xi_ep_fig}
\end{figure*}

To ensure that the object samples are gradually involved from easy to hard as the training proceeds, the value of $\xi$ should decrease gradually with increasing training iterations. In this paper, we design an example of the relationship between $\xi$ and the training process, as shown in Fig. \ref{xi_ep_fig}. The interpretation of the relationship between $\xi$ and training process as follows: when the training starts (the training progress is less than or equal to $e_1$), the flying bird objects with prediction confidence greater than $\xi_0$ will participate in the training, and the rest of the objects will not participate in the training. When the training progress is between $e_1$ and $e_2$, the confidence threshold decreases linearly, and the hard objects gradually participate in the training. At the end of the training (the training progress reaches more than $e_2$), all objects participate in the training (in the subsequent experiments, $\xi_0$ is set to 0.8, $e_1$ and $e_2$ are set to 10\% and 90\%, respectively).

The relationship between the confidence threshold parameter $\xi$ and the training progress EP can be expressed as,
\begin{align}\label{eq:xi_ep}
\xi = 
\begin{cases}
\xi_0, &\text{EP} < e_1,\\
\frac{\xi_0\left( e_2 - \text{EP} \right)}{e_2 - e_1}, &e_1 \leq \text{EP} < e_2,\\
0, &e_2 \leq \text{EP}.
\end{cases}
\end{align}
From the perspective of curriculum learning, Formula \eqref{eq:mf_example} is the difficulty measurement method of the object sample of flying birds, Formula \eqref{eq:xi_ep} is the method of training scheduling, and $\xi$ is the training scheduling parameter.

\subsection{The FBOD Model Training strategy based on SPL-ESP-BC}\label{step2}

Before applying the SPL-BC strategy to train the FBOD model, this paper employed easy samples for pre-training, thus preventing the model from falling into a disordered search state due to its initial inability to discriminate between easy and hard samples. The aforementioned training strategy is referred to as the Self-Paced Learning strategy with Easy Sample Priority Based on Confidence (SPL-ESP-BC). Next, we will describe the FBOD model training strategy based on SPL-ESP-BC in detail.

Some easy object samples are selected manually to train the FBOD model so that it has the ability to identify the difficulty of the sample initially. Specifically, the flying bird objects are divided into easy and hard objects by whether they can be easily identified. Select a part of easy samples ($S_e \subset S$, where $S$ represents the set of all flying bird object samples and $S_e$ represents the set of easy object samples selected manually). Due to subjective factors, different people choose different sets of $S_e$. Some hard samples will be included, but most will be easy samples. Due to the robustness of the deep learning model, a small number of hard samples will not significantly affect the learning process of the FBOD model. Then, the manually selected easy sample set $S_e$ is used to train the FBOD model $f=\mathcal{N}_\text{FBOD}$, and the weight parameter $\mathbf{w}=\bf{\Theta}_\text{FBOD}^{'}$ of the model is obtained. This weight parameter is used as the initial weight parameter of the model in the subsequent SPL process (when the model is trained with easy samples, its weight parameter was randomly initialized with simple Gaussian).

After training the FBOD model with easy samples, it can preliminarily identify easy and hard samples. In the subsequent training, we use all the samples (dataset $S$) to train the FBOD model. Specifically, firstly, $\bf{\Theta}_\text{FBOD}^{'}$ is used to initialize the weight $\mathbf{w}$ of the FBOD model $\mathcal{N}_\text{FBOD}$, and then all the bird samples (dataset $S$) are used to train the FBOD model by the SPL-BC. In alternating iterative optimization, the model weight is optimized by the Weighted Loss Function shown in Eq. \eqref{eq:wlf}, and sample weights are optimized using the Minimizer Function based on confidence shown in Eq. \eqref{eq:mf_example}.

The pseudo-code of the training strategy of the FBOD model based on SPL-ESP-BC is shown in Algorithm \ref{algo1}.
\begin{algorithm}
\caption{The training strategy of the FBOD model based on SPL-ESP-BC}\label{algo1}
\begin{algorithmic}[1]
\Require Flying birds dataset $S$, FBOD model $f=\mathcal{N}_\text{FBOD}$, the number of iterations $T$;
\Ensure FBOD model with weight parameters $\mathbf{w}=\bf{\Theta}_\text{FBOD}$.
\State Let $T=T_0 + T_1$;
\State Select easy dataset $S_e \subset S$;
\State Initialize $\mathbf{w}$ with simple gaussian, initialize $t = 0$;
\While{$t \neq T_0 $}
        \State $t = t + 1$;
        \State Select a batch of images and corresponding labels from $S_e$ randomly;
        \State Input the images into the FBOD model to get outputs;
        \State Input the outputs and labels into Eq. \eqref{eq:tl}, update $\mathbf{w}$ by gradient descent;
\EndWhile
\State Freeze $\mathbf{w} = \bf{\Theta}_\text{FBOD}^{'}$;
\State Initialize $\mathbf{w}  = \bf{\Theta}_\text{FBOD}^{'}$, $\boldsymbol{\Vec{v}} = \boldsymbol{0}$, $\xi = \xi_0$, $t = 0$;
\While{$t \neq T_1 $}
        \State $t = t + 1$;
        \State Select a batch of images and corresponding labels from $S$ randomly;
        \State Fix $\mathbf{w}$, input the images into the FBOD model to get the outputs;
        \State Input the outputs into Eq. \eqref{eq:mf_example} to update $\boldsymbol{\Vec{v}}$;
        \State Fix $\boldsymbol{\Vec{v}}$, input the outputs and labels into Eq. \eqref{eq:wlf}, update $\mathbf{w}$ by gradient descent;
        \State Update $\xi$ through Eq. \eqref{eq:xi_ep});// To include more hard samples
\EndWhile
\State Freeze $\mathbf{w} = \bf{\Theta}_\text{FBOD}$.
\end{algorithmic}
\end{algorithm}

 \section{Experiments}\label{experiments}

 In this part, quantitative and qualitative experiments will be conducted to demonstrate the effectiveness and advancement of the proposed method. Next, the dataset, evaluation method, implementation details, and comparative analysis experiments will be introduced.

 \subsection{Datasets}\label{datestes}

 The dataset used to verify this method is consistent with the dataset in our last paper \cite{sun2024flying}. The dataset has 120 videos containing flying bird objects with 28,353 images. Among them, 101 videos (24898 images) are used as the training set, and 19 videos (3455 images) are used as the test set. Refer to \cite{sun2024flying} for more details on this dataset.

 \subsection{Evaluation Metrics}\label{evaluation_metrics}

 In this paper, referring to the evaluation indexes of other object detection algorithms, the average precision (AP) evaluation index of Pascal VOC 2007 \cite{2010_Pascal_VOC} is used to evaluate the detection results of the model.

 \subsection{Implementation Details}\label{implementation_details}

 The FBOD model \cite{sun2024flying} we designed before will be used in this paper. The input of the model is five consecutive 3-channel RGB images of size 672×384, and the output is 336×192×1 confidence prediction feature map and 336×192×4 position regression feature map. The output predicts the position of the flying bird object on the intermediate frame.
 
In this paper, all experiments are implemented under the Pytorch framework. The network models are trained on an NVIDIA GeForce RTX 3090 with 24 GB of video memory. All experimental models are trained from scratch without pre-trained models. The trainable parameters of its network are randomly initialized using a normal distribution with a mean of 0 and a variance of 0.01. Adam was chosen as the optimizer for the model in this paper. The initial learning rate is set to 0.001, and for each iteration, the learning rate is multiplied by 0.95, and the model is trained for 150 iterations. Among them, In these training strategies of SPL with ESP, the ESP stage has 50 iterations ($T_0 = 50$ in algorithm \ref{algo1}), and the SPL stage has 100 iterations ($T_1 = 100$ in algorithm \ref{algo1}). When the model is trained, batch size is set to 8.

\subsection{Comparative Analysis Experiments}\label{comparative}

To prove that the proposed method is effective and advanced, we set up two sets of comparative experiments related to it.

The first set of comparative analysis experiments will verify the effectiveness of the proposed method by comparing four training modes of the model. The four training modes are Easy Sample (ES), All Sample (AS), Hard Example Mining (HEM), and Self-Paced Learning training strategy with Easy Sample Prior Based on Confidence (SPL-ESP-BC). Among them, for the ES training mode, the hard samples in the dataset are manually selected and eliminated, and only the easy samples are trained by the ordinary training method (random gradient descent method, without distinguishing the sample conditions). For the AS training mode, all samples use ordinary training methods to train the model. For the HEM training mode, the loss of all samples is calculated before each iteration of training, and the loss is sorted from major to small. The first 40\% of samples are selected, and ordinary training methods train the model. For the SPL-ESP-BC training mode, easy samples are used to train the model with ordinary training methods first, and then, using all the samples, the SPL-BC strategy is used to train the model. In this paper, the first set of comparative analysis experiments is called the comparative experiment of different training modes.

The second set of comparative analysis experiments will verify the advancement of the proposed method by comparing four different SPL strategies. The four SPL strategies are SPL strategy with Easy Sample Prior based on Hard regularizer \cite{kumar_SPL_for_latent_variable} (SPL-ESP-BH), SPL strategy with ESP based on Linear regularizer \cite{jiang_easy_samples_first_sp_regularizers} (SPL-ESP-BLine), SPL strategy with ESP based on Logarithmic regularizer \cite{jiang_easy_samples_first_sp_regularizers} (SPL-ESP-BLog) and the SPL-ESP-BC strategy proposed in this paper. SP-ESP-BH, SP-ESP-BLine, and SP-ESP-blog are SPL strategies based on sample loss compared with the methods proposed in this paper. In this paper, the second set of comparative analysis experiments is called the comparative experiment of different SPL strategies.

\subsubsection{The Comparative Experiment of Different Training Modes}

Table \ref{tab:4_modes_detection_accuracy} shows the quantitative evaluation results of the four training modes. The results show that the FBOD model's $\text{AP}_{50}$ trained by the SPL-ESP-BC mode reaches 0.782, which is 2.1\% higher than that of the AS training mode, 6.1\% higher than that of the ES training mode, and 5.8\% higher than that of the HEM mode. The results confirm that the proposed method achieves the best results. For AP75 and AP, the proposed model training mode SPL-ESP-BC also greatly improved.

\begin{table}[!ht]
\caption{Detection accuracy of FBOD models trained by four different training modes.\label{tab:4_modes_detection_accuracy}}
\centering
\begin{tabular}{c|c c c}
\hline
Mode  & $\text{AP}_{50}$ & $\text{AP}_{75}$  & AP\\
\hline
AS  & 0.762 & 0.371  & 0.395\\
ES  & 0.722 & 0.304  & 0.345\\
HEM  & 0.725 & 0.211  & 0.310\\
SPL-ESP-BC  & 0.782 & 0.369  & 0.398\\
\hline
\end{tabular}
\end{table}

The ES training mode, which uses only easy samples to train the model, makes it difficult to detect hard samples in the test set in the test phase. Compared with the ES training mode, the AS training mode increases the hard samples in the training process. In the test stage, the easy and hard samples in the test set can be detected. The HEM training mode, which uses only hard samples to train the model, tends to make the model overfit the hard samples. The SPL-ESP-BC training mode first uses easy samples to train the model and then gradually introduces hard samples, which can suppress the noise caused by hard samples to a certain extent. Therefore, the FBOD model trained by the training mode proposed in this paper achieves the highest detection accuracy.

To further analyze the difference in the detection performance of the FBOD model trained by different training modes, we calculate the detection rate of bird objects in each difficulty level and the false detection rate in the whole test set. Specifically, we first manually divide the difficulty level of the flying bird objects in the test set (manual division, there is a certain degree of subjectivity, but it does not affect the relativity of the degree of difficulty, so different training modes can be judged), and divide into four levels, namely difficulty level 1, ..., difficulty level 4, where the higher the difficulty level, the harder the sample is to identify. Then, the detection rate of the FBOD model trained by different training modes in each difficulty level and the false detection rate in the whole test set is calculated.

The statistical results are shown in Table \ref{tab:4_modes_Detection_rate}. As can be seen from Table \ref{tab:4_modes_Detection_rate}, the FBOD model trained by the ES training mode can detect easy samples well, but the detection rate of hard samples is low. The FBOD model trained by the AS training mode can not only detect the easy samples, but also the hard samples have a high detection rate, but the false detection rate is also high, which shows that simply adding the hard samples will cause some noise influence. The FBOD model trained by HEM training mode has a higher detection rate for objects of different difficulty levels, but its false detection rate is also much higher than the other three modes, which indicates that only the hard samples will make the model overfit the hard samples, resulting in more false detection. The FBOD model trained by the SPL-ESP-BC training mode has a high detection rate for hard samples and keeps the false detection rate low. Therefore, using the proposed method to train the FBOD model can improve the detection rate of hard samples and suppress the noise caused by hard samples.
\begin{table}[!ht]
\caption{The false detection rate of the FBOD models trained by four different training modes and the detection rate of samples with different difficulty levels.\label{tab:4_modes_Detection_rate}}
\centering
\begin{tabular}{c|c c c c c}
\hline
Mode  & \makecell[c]{Difficulty\\Level 1} & \makecell[c]{Difficulty\\Level 2} & \makecell[c]{Difficulty\\Level 3} & \makecell[c]{Difficulty\\Level 4} & \makecell[c]{False\\Detection}\\
\hline
AS  & 0.993 & 0.798  & 0.711 & 0.749 & 0.131\\
ES  & 0.991 & 0.677  & 0.452 & 0.226 & 0.045\\
HEM  & 0.998 & 0.784  & 0.760 & 0.829 & 0.173\\
SPL-ESP-BC  & 0.993 & 0.781  & 0.718 & 0.865 & 0.053\\
\hline
\end{tabular}
\end{table}

\begin{figure*}[!ht]
\centering
\includegraphics[width=5in]{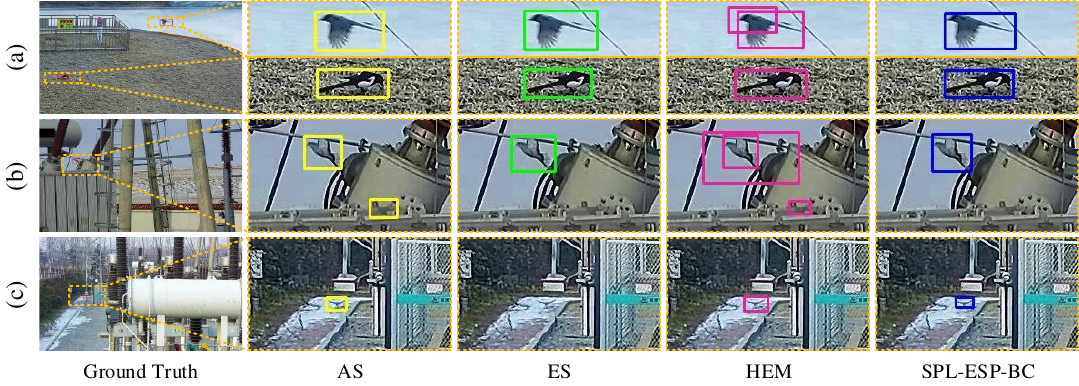}
\caption{The detection effects of FBOD models are trained using four different training modes in three different situations.}
\label{results_fig}
\end{figure*}

Fig. \ref{results_fig} shows the detection effects of the FBOD model trained by four training modes in three scenarios. Among them, the bird object in scene 1 is relatively easy to identify, the bird object in scene 2 is slightly difficult, and the bird object in scene 3 is difficult to identify. For easy samples, models trained by the four training modes can all be well detected, as shown in Fig. \ref{results_fig}(a). The AS training mode simply and directly adds hard samples, which will cause certain noise effects, and the trained model is more prone to false detection, as shown in Fig. \ref{results_fig}(b). The model trained by ES training mode has a poor detection effect on hard samples because it does not use hard samples to train the model, as shown in Fig. \ref{results_fig}(c). No matter whether the bird objects are easy or not, the model trained by HEM mode can detect them, but the false detection is also serious, as shown in Fig. \ref{results_fig}(a), \ref{results_fig}(b), and \ref{results_fig}(c). However, the model trained by the SPL-ESP-BC mode in this paper can detect samples of different difficulty levels better and have fewer false detection cases. The visualization results further prove the above view, that is, the FBOD model trained by SPL-ESP-BC training mode not only has a high detection rate for hard samples but also keeps its false detection rate low.

\subsubsection{The Comparative Experiment of Different SPL Strategies}

\begin{figure*}[!ht]
\centering
\includegraphics[width=2.5in]{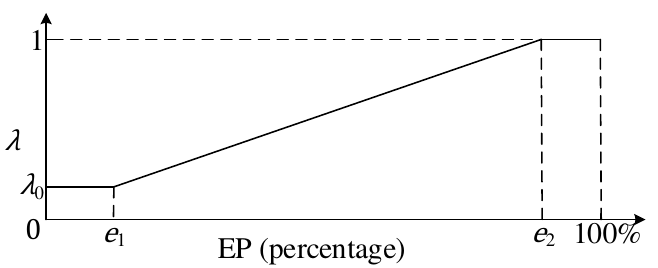}
\caption{The relationship between $\lambda$ and training process.}
\label{lambda_ep_fig}
\end{figure*}

The SPL strategy proposed in this paper is more suitable than the loss-based SPL strategy when applied to one-class object detection. We set up a contrastive analysis experiment of different SPL strategies to prove this point. Specifically, SPL strategies based on Hard \cite{kumar_SPL_for_latent_variable}, Linear \cite{jiang_easy_samples_first_sp_regularizers}, and Logarithmic \cite{jiang_easy_samples_first_sp_regularizers} regularizers are incorporated to compare with the proposed strategy in this paper. To simplify the process of solving the optimal weight (the analytic expression of its closed solution is shown in Table \ref{tab:Minimizer_Functions}), we adopted a certain method to make the sample loss\footnote{This sample loss is only used to solve the optimal weight of the sample in SPL. The objective function of the optimization model's weight still adopts the Weighted Loss Function shown in Eq. \eqref{eq:wlf}, in which the sample loss involved is calculated in the same way as the method proposed in this paper.} of calculating the optimal weight between (0,1).Specifically, if the bird sample $F_i$ has $K$ anchor point samples, then the loss of the bird sample is,
\begin{align}\label{eq:lak}
l_{A_k} =\frac{1}{2}\left(\Vert \text{Conf}_{\text{pred}}\left(A_k\right) - 1 \Vert + \frac{1}{2}\text{LCIOU}\left( \text{box}_\text{gt}\left(A_k\right), \text{box}_\text{pred}\left(A_k\right) \right) \right),
\end{align}
\begin{align}\label{eq:lfi}
l_{F_i} =\frac{1}{K}\sum\limits_{k=0}^K l_{A_k},
\end{align}
where $l_{A_k}$ represents the loss of the $k^\text{th}$ anchor sample of the flying bird sample $F_i$. $\text{LCIOU}(\cdot)$ is the CIOU loss, whose value range is (0,2), and $l_{F_i}$ is the sample loss of the bird sample $F_i$, whose range is (0,1). For the age parameter $\lambda$ used for scheduling training, a similar adjustment strategy as the training scheduling parameter $\xi$ in this paper is adopted (where $\lambda_0$ is set to 0.2, $e_1$ and $e_2$ are set to 10\% and 90\%, respectively) as follows,
\begin{align}\label{eq:lambda_ep}
\lambda = 
\begin{cases}
\lambda_0, &\text{EP} < e_1,\\
\frac{\left( 1 - \lambda_0 \right)}{e_2 - e_1} \left( e_2-\text{EP} \right) + 1, &e_1 \leq \text{EP} < e_2,\\
1, &e_2 \leq \text{EP}.
\end{cases}
\end{align}
Fig. \ref{lambda_ep_fig} demonstrates the relationship between the age parameter $\lambda$ and the training progress EP. Based on Fig. \ref{lambda_ep_fig}, the explanation of this training schedule is as follows: in the initial stage of training (training progress $\text{EP} < e_1$), only the samples with loss less than $\lambda_0$ participate in the training; when the training process is between $e_1$ and $e_2$, the threshold of sample loss gradually increases, and hard samples gradually participate in the training; in the final stage of training (training progress $\text{EP} \geq e_2$), all samples participate in the training. The optimal weights of corresponding samples can be obtained by substituting Eq. \eqref{eq:lfi} and \eqref{eq:lambda_ep} into the analytical solutions of optimal weights closed-form for three types of regularizers in Table \ref{tab:Minimizer_Functions}. These weights control whether the corresponding samples participate in the training or the degree of their participation.

To ensure consistency with other conditions of the method proposed in this paper, the model is also trained by ordinary training methods using easy samples (Easy Sample Prior) before the other three SPL strategies are used.

Tables \ref{tab:4_strategies_detection_accuracy} and \ref{tab:4_strategies_Detection_rate} show the evaluation results of the FBOD models trained by different SPL strategies on the test set. Those marked in red are \textcolor[rgb]{1,0,0}{optimal}, and those marked in green are \textcolor[rgb]{0,1,0}{suboptimal}. The table shows that although the proposed method is not optimal in every index, its comprehensive performance is optimal from the perspective of optimal and suboptimal. Therefore, the SPL-ESP-BC strategy proposed in this paper is advanced and more suitable for bird object detection model in surveillance video.
\begin{table}[!ht]
\caption{Detection accuracy of FBOD models trained by four different training strategies.\label{tab:4_strategies_detection_accuracy}}
\centering
\begin{tabular}{c|c c c}
\hline
Strategy  & $\text{AP}_{50}$ & $\text{AP}_{75}$  & AP\\
\hline
SPL-ESP-BH \cite{kumar_SPL_for_latent_variable}  & 0.771 & \textcolor[rgb]{1,0,0}{0.372}  & 0.385\\
SPL-ESP-BLine \cite{jiang_easy_samples_first_sp_regularizers}  & \textcolor[rgb]{1,0,0}{0.783} & 0.346  & \textcolor[rgb]{0,1,0}{0.389}\\
SPL-ESP-BLog \cite{jiang_easy_samples_first_sp_regularizers}  & 0.743 & 0.367  & 0.379\\
SPL-ESP-BC(this paper)  & \textcolor[rgb]{0,1,0}{0.782} & \textcolor[rgb]{0,1,0}{0.369}  & \textcolor[rgb]{1,0,0}{0.398}\\
\hline
\end{tabular}
\end{table}

\begin{table}[!ht]
\caption{The false detection rate of the FBOD models trained by four different training strategies and the detection rate of samples with different difficulty levels.\label{tab:4_strategies_Detection_rate}}
\centering
\begin{tabular}{c|c c c c c}
\hline
Strategy  & \makecell[c]{Difficulty\\Level 1} & \makecell[c]{Difficulty\\Level 2} & \makecell[c]{Difficulty\\Level 3} & \makecell[c]{Difficulty\\Level 4} & \makecell[c]{False\\Detection}\\
\hline
SPL-ESP-BH \cite{kumar_SPL_for_latent_variable}  & \textcolor[rgb]{1,0,0}{0.995} & 0.779  & \textcolor[rgb]{1,0,0}{0.724} & 0.802 & 0.063\\
SPL-ESP-BLine \cite{jiang_easy_samples_first_sp_regularizers}  & \textcolor[rgb]{0,1,0}{0.993} & \textcolor[rgb]{1,0,0}{0.801}  & 0.714 & \textcolor[rgb]{0,1,0}{0.863} & \textcolor[rgb]{0,1,0}{0.056}\\
SPL-ESP-BLog \cite{jiang_easy_samples_first_sp_regularizers}  & \textcolor[rgb]{1,0,0}{0.995} & 0.776  & 0.706 & 0.767 & 0.090\\
SPL-ESP-BC(this paper)  & \textcolor[rgb]{0,1,0}{0.993} & \textcolor[rgb]{0,1,0}{0.781}  & \textcolor[rgb]{0,1,0}{0.718} & \textcolor[rgb]{1,0,0}{0.865} & \textcolor[rgb]{1,0,0}{0.053}\\
\hline
\end{tabular}
\end{table}

\section{Conclusion}\label{conclusion}

This paper proposes a new training strategy called Self-Paced Learning strategy with Easy Sample Prior Based on Confidence (SPL-ESP-BC). Firstly, the loss-based Minimizer Function in Self-Paced Learning (SPL) is improved, and a confidence-based Minimizer Function is proposed, which makes it more suitable for one-class object detection tasks. Secondly, an SPL strategy with Easy Sample Prior (ESP) is proposed. The FBOD model is trained with easy samples by using the ordinary training method first, and then the model training strategy of SPL is adopted, and all samples are used to train it. In this way, the model has the ability to judge easy samples and hard samples in the early stage of the SPL strategy. Finally, the SP-ESP-BC strategy is proposed by combining the ESP strategy with the confidence-based Minimizer Function. The SPL-ESP-BC strategy is used to train the FBOD model, which can make it better learn the characteristics of flying birds in surveillance videos from easy to hard. Through experimental verification, it is proved that the model training strategy proposed in this paper is effective and advanced.

 \bibliographystyle{elsarticle-num} 
 \bibliography{cas-refs}





\end{document}